%% file: main.tex
\documentclass{article}
\input{include/packages}

\icmltitlerunning{Heuristic Search as Language-Guided Program Optimization}

\begin{document}

\twocolumn[
  \icmltitle{Heuristic Search as Language-Guided Program Optimization}

  \icmlsetsymbol{equal}{*}

  \begin{icmlauthorlist}
    \icmlauthor{Mingxin Yu}{mitaa}
    \icmlauthor{Ruixiao Yang}{mitaa}
    \icmlauthor{Chuchu Fan}{mitaa}
  \end{icmlauthorlist}

  \icmlaffiliation{mitaa}{Massachusetts Institute of Technology, Cambridge, MA, USA}

  \icmlcorrespondingauthor{Mingxin Yu}{yumx35@mit.edu}

  \vskip 0.3in
]

\printAffiliationsAndNotice{}  %

\begin{abstract}

\input{tex/00-abstract}
\end{abstract}

\section{Introduction}\input{tex/01-introduction}

\section{Related Works}\input{tex/03-related}

\section{Method}
\input{tex/05-formulation}
\input{tex/06-method}

\section{Experiment}\input{tex/07-exp}

\section{Conclusion and Discussion}\input{tex/09-conclusion}

\newpage
\section*{Impact Statement}
This paper presents work whose goal is to advance the field of Machine
Learning. There are many potential societal consequences of our work, none
which we feel must be specifically highlighted here.

\bibliography{tex/ref}
\bibliographystyle{icml2026}

\newpage
\appendix
\onecolumn %

\section{Overview of Appendix}
This Appendix contains several sections, each addressing a specific aspect of the experimental setups and results. Below is a brief overview of each section:
\begin{itemize} 
    \item Discussion on additional related works of executable feedback in code generation;
    \item More details about the problems we include in \cref{tab: original-baseline}, and the reasons we select them from the benchmark while excluding the rest;
    \item Additional experiment results using \texttt{GPT-4.1-mini} on PDPTW;
    \item The prompts we use in our system.
\end{itemize}

\section{Additional Related Works}\input{appendix/related}

\section{Combinatorial Problems in HeuriGym}\input{appendix/heurigym}

\section{Experiment}\input{appendix/more-results}

\section{Prompt Design}~\input{appendix/prompt}

\end{document}

%% file: include/packages.tex
\usepackage{microtype}
\usepackage{graphicx}
\usepackage{subcaption}
\usepackage{booktabs} %

\usepackage{hyperref}

\usepackage[preprint]{icml2026}

\usepackage{amsmath}
\usepackage{amssymb}
\usepackage{mathtools}
\usepackage{amsthm}

\usepackage[capitalize,noabbrev]{cleveref}

\theoremstyle{plain}

\theoremstyle{definition}

\theoremstyle{remark}

\usepackage[textsize=tiny]{todonotes}

\usepackage{multirow}
\usepackage{nicematrix, booktabs}
\usepackage{makecell}
\usepackage[most]{tcolorbox}

%% file: tex/00-abstract.tex
Large Language Models (LLMs) have advanced Automated Heuristic Design (AHD) in combinatorial optimization (CO) in the past few years. 
However, existing discovery pipelines often require extensive manual trial-and-error or reliance on domain expertise to adapt to new or complex problems. 
This stems from tightly coupled internal mechanisms that limit systematic improvement of the LLM-driven design process.
To address this challenge, we propose a structured framework for LLM-driven AHD that explicitly decomposes the heuristic discovery process into modular stages: a forward pass for evaluation, a backward pass for analytical feedback, and an update step for program refinement. This separation provides a clear abstraction for iterative refinement and enables principled improvements of individual components. 
We validate our framework across four diverse real-world CO domains, where it consistently outperforms baselines, achieving up to $0.17$ improvement in QYI on unseen test sets. 
Finally, we show that several popular AHD methods are restricted instantiations of our framework. By integrating them in our structured pipeline, we can upgrade the components modularly and significantly improve their performance.

%% file: tex/01-introduction.tex
Heuristics play a critical role in solving complex search and optimization problems, where exact algorithms are often computationally challenging~\cite{lin1973lkh,lourencco2003iterated,hromkovivc2013algorithmics}. Automated Heuristic Design (AHD) aims to reduce the substantial manual effort and domain expertise required to design such heuristics by automatically discovering effective strategies~\cite{Burke2013}.
However, traditional AHD methods often rely on hand-crafted algorithmic primitives, which restrict their search space and limit their ability to generalize to novel problem settings~\cite{o2010open,stutzle2018automated,drake2020recent}.

Recent advancements in Large Language Models (LLMs) have significantly expanded the scope of AHD by enabling the direct generation of executable code from natural language descriptions or partial program specifications~\cite{austin2021program,chen2021evaluating}. Unlike traditional methods that operate within a fixed, predefined search space, LLM-based approaches explore a far more open-ended space of heuristic programs~\cite{chen2023evoprompting,meyerson2024language}. 
A growing body of work demonstrates that LLMs can discover competitive heuristics for challenging CO problems by iteratively proposing, evaluating, and refining heuristic programs~\cite{romera2024funsearch,fei2024eoh}.

Recent studies have proposed various enhancements for LLM-based AHD, including refined prompt engineering or specialized feedback mechanisms~\cite{ye2024reevo,ye2025llmlns, cui2025heuristic}. Despite empirical successes, the development of these systems remains driven by manual trial-and-error or expert insights.
This is largely because these advancements are typically introduced within tightly coupled heuristic discovery pipelines, where evaluation, feedback, and update logic are implemented as a single, intertwined loop.
Such coupling makes it difficult to isolate the contribution of individual components or systematically improve the LLM-driven heuristic design process and to transfer effective design strategies across new or complex problem domains.

In this work, we address these challenges by reformulating AHD as an explicit optimization problem over a discrete program space. We propose \textbf{La}nguage-\textbf{G}uided \textbf{O}ptimization (\textbf{LaGO}), a framework decomposing the heuristic discovery process into three distinct, modular stages: 
(1) a forward pass evaluating candidate heuristic and records execution-level behavior; 
(2) a backward pass analyzing behaviors to derive structured feedback; and 
(3) an update stage refining the heuristic code based on the feedback. 
This decomposition transforms the search from a stochastic mutation process into a directed optimization. By disentangling these components, LaGO enables principled, "plug-and-play" upgrades to individual stages of the search process.

Our contributions are summarized as follows:
\begin{itemize}
    \item We introduce LaGO, a unified framework that formalizes LLM-based AHD as a language-guided optimization problem. By decomposing the pipeline into forward, backward, and update modules, LaGO enables systematic analysis of the heuristic search process and principled modification of its components.
    \item LaGO provides a general mechanism for improving existing AHD methods by re-implementing prior approaches as LaGO instantiations and upgrading individual modules. Specifically, we introduce executable feedback in the backward pass, a joint optimization strategy to improve constructive and refinement heuristics, and a diversity-aware population management strategy for the update stage.
    \item We validate LaGO across four real-world CO case studies with challenging feasibility constraints. Across all domains, LaGO consistently outperforms existing baselines, while detailed analysis shows that the proposed modular upgrades improve search efficiency and help avoid common failure modes like mode collapse.
\end{itemize}

%% file: tex/03-related.tex
\subsection{Automated Heuristic Design (Pre-LLM)}
The automation of algorithm design, historically referred to as hyper-heuristic~\cite{Burke2013}, aims to reduce the manual effort required to solve combinatorial optimization problems. Early methods~\cite{hutter2009paramils, blot2016moparamils, lopez2016irace, akiba2019optuna} focus primarily on selecting, combining, and tuning predefined algorithmic components in an automated manner. 
Beyond simple selection, evolutionary approaches such as genetic programming~\cite{mei2022explainable, xu2023genetic} have also been used to automatically synthesize heuristics and algorithmic components. These methods can yield interpretable, domain-specific heuristics and have shown practical success in real-world applications~\cite{zhu2023surgical,zhang2023qjssp}.
However, these approaches still require substantial human effort to design the search space. This requirement for domain expertise limits flexibility and can constrain further performance improvements.

\subsection{LLM-Based Heuristic and Algorithm Discovery}
The emergence of Large Language Models (LLMs) has introduced a new paradigm in algorithm discovery by enabling the direct synthesis of executable algorithms~\cite{yang2023llmassolver, lange2024llmases}. Early approaches primarily use LLMs as black-box solvers, using in-context learning to generate solutions for individual problem instances. While effective for small or structured instances, this solver-style paradigm often suffers from limited generalization, particularly when scaling to complex problems or larger instance sizes.
More recent work has shifted toward iterative heuristic evolution, where LLMs are used to generate and refine algorithmic components through repeated evaluation and selection~\cite{romera2024funsearch, fei2024eoh} or Monte-Carlo tree search~\cite{zheng2025mcts}. These approaches introduce an explicit evaluation loop that enables the discovery of reusable heuristics, and have been successfully adapted to domain-specific settings such as large neighborhood search~\cite{ye2025llmlns}. 

Despite their empirical success~\cite{dat2025hsevo,zheng2025mcts, zhong2025llmoa}, most existing methods operate as forms of randomized evolutionary search: LLMs are typically employed as stochastic mutation or crossover operators guided by scalar fitness values, with limited insight into why a heuristic succeeds or fails.
As a result, the optimization process in these methods remains largely implicit. Evaluation and update mechanisms are tightly coupled, and feedback is restricted to coarse performance signals, making it difficult to systematically analyze, compare, or extend the underlying search dynamics. ReEvo~\cite{ye2024reevo} and CALM~\cite{huang2025calm} partially addresses this limitation by introducing textual reflection into the evolutionary process, but the overall evolution process remains entangled.
In contrast, our work focuses on making the optimization structure of LLM-based heuristic discovery explicit. By formulating heuristic design as an optimization problem over a discrete program space and decomposing the process into stages, we enable systematic analysis and principled improvement of heuristic discovery methods, moving beyond ad-hoc evolutionary search.

\begin{figure*}[h]
    \centering
    \includegraphics[width=1\linewidth]{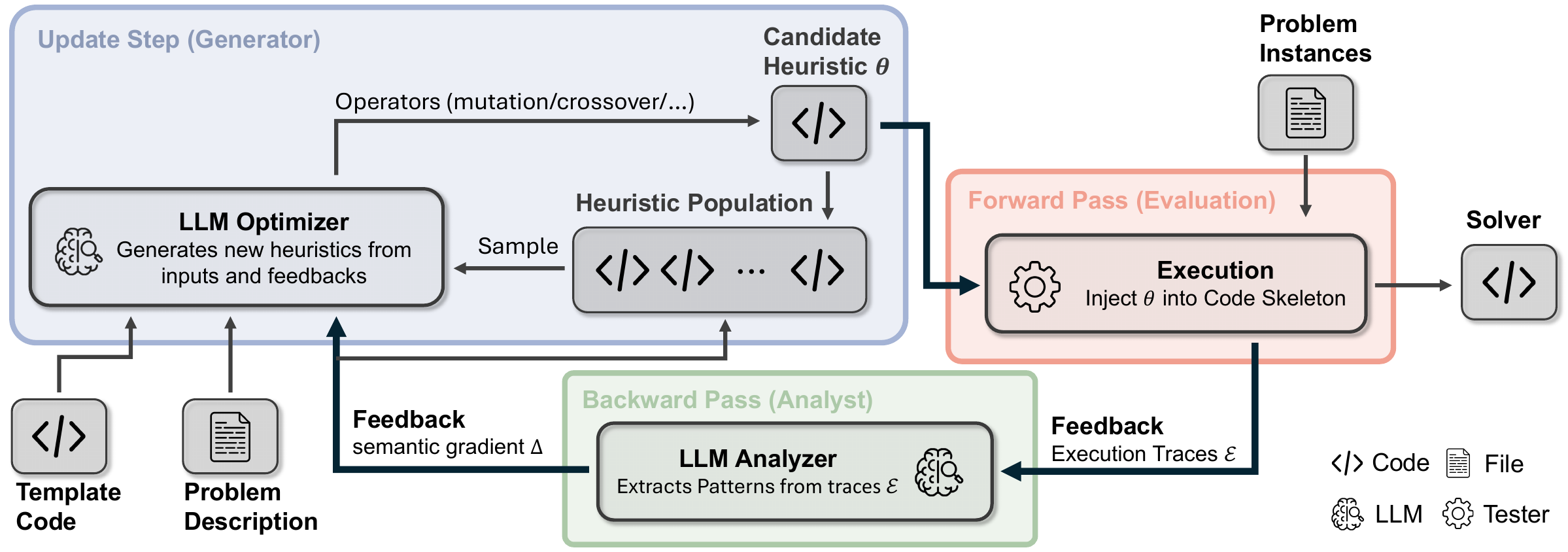}
    \caption{\textbf{The LaGO framework for language-guided optimization.} Our framework decomposes the automated heuristic design process into three modular stages: a forward pass (red), a backward pass (green), and an update step (blue).
This modular structure supports systematic refinement of heuristic logic while preserving compatibility across different problem domains.}
    \label{fig:pipeline}
\end{figure*}

%% file: tex/05-formulation.tex
We propose LaGO, a modular framework that formalizes Automated Heuristic Design (AHD) as a structured optimization process. The entire framework is illustrated in~\cref{fig:pipeline}. In the following sections, we will detail our problem formulation and the specific module architectures that instantiate this language-guided optimization pipeline.

\subsection{Problem Formulation: Heuristic Search as Program Optimization}
We formally define the Heuristic Design Problem as an optimization task over a space of executable programs $\Theta$. Consider a distribution of Combinatorial Optimization (CO) problem instances $\mathcal{D}_{target}$. Each problem instance is defined as a tuple $\Phi = (\mathcal{X}, g, \mathcal{C}, \mathcal{T})$, where $\mathcal{X}$ denotes the decision variable space of the combinatorial problem, $g: \mathcal{X} \mapsto \mathbb{R}^+\cup\{0\}$ is the objective function to be minimized, $\mathcal{C}$ is the set of hard constraints, which a valid solution $x \in \mathcal{X}$ must satisfy, and $\mathcal{T}$ is a natural language description of the problem logic and constraints, which provides the semantic context required for Large Language Models (LLMs).

Following recent works~\cite{romera2024funsearch,fei2024eoh} in algorithm discovery, we search for a heuristic policy $h_\theta$, parameterized by executable code $\theta \in \Theta$, that constructs or improves a solution $x = h_\theta(\Phi)$. The quality of a heuristic is measured by a fitness metric 
\begin{equation}
    f(\Phi, x)=\min\Big\{1, \frac{g(x^*)}{g(x)}\Big\} \in [0, 1],
\end{equation} 
where $x^*$ is the best-known solution, or an optimal solution if available. Our objective is to find the optimal program parameters $\theta^*$ that maximize the expected fitness over the target distribution:
\begin{equation}
    \theta^* = \arg\max_{\theta \in \Theta} \mathbb{E}_{\Phi \sim \mathcal{D}_{target}} \left[ f(\Phi, h_\theta(\Phi)) \right]
    \label{eq:problem}
\end{equation}
In practice, $\mathcal{D}_{target}$ is unknown. We are provided with a limited training set $\mathcal{D}_{train} = \{\Phi_1, \dots, \Phi_K\}$. The goal is to learn a heuristic $\hat{\theta}^*$ on $\mathcal{D}_{train}$ that generalizes to a larger, potentially out-of-distribution test set $\mathcal{D}_{test}$.

\subsection{The LaGO Framework: Language-Guided Optimization}

In order to solve the problem in~\cref{eq:problem}, we propose a general framework, \textbf{LaGO}. As illustrated in~\cref{fig:pipeline}, our framework reformulates the heuristic evolution process by drawing a direct parallel to gradient-based learning \cite{lecun2015deep}. The system decomposes the optimization process into three stages: a \textbf{Forward Pass} (evaluation), a \textbf{Backward Pass} (analyst), and an \textbf{Update Step} (generator).

Following~\cite{fei2024eoh, dat2025hsevo, ye2024reevo}, we maintain a population of candidate heuristics across optimization iterations. Formally, we define the optimization step at iteration $t$ as an update to a population of program parameters $P_t = \{\theta_t^{(i)}\}_{i=1}^N$ to next population $P_{t+1}$. The framework is composed of the following components:

\paragraph{Forward Pass (Evaluation).} The forward pass takes a heuristic $\theta$ as input, injects it into the provided algorithm skeleton, and evaluates it on the training set $\mathcal D_{train}$. The process yields an execution trace  $\mathcal E_\theta$
containing runtime errors, generated solutions, and potentially other intermediate messages. These traces then provide the raw empirical data for the feedback signal in the backward pass.

\paragraph{Backward Pass (Analyst).}
To guide the population update, we require a feedback signal analogous to the gradient $\nabla_\theta J$. Standard evolutionary methods constrict this signal to a scalar fitness value or plain text, causing significant information loss. We introduce the \textbf{Analyst} module $\mathcal A$, which maps execution traces to a \textbf{semantic gradient} $\Delta$:
\begin{equation}
    \Delta_t^{(i)} \leftarrow \mathcal{A}(\mathcal E_t^{(i)}, \mathcal{T}).
\end{equation}
Crucially, LaGO imposes no constraints on the format of $\Delta$. It serves as a flexible container for the information critical to improvement. Depending on the domain, $\Delta$ may be:
\begin{itemize}
    \item \textbf{Structured Tensors:} High-dimensional vectors concatenating fitness with other statistics.
    \item \textbf{Unstructured Semantics:} Natural language critiques, debug logs, or code derived from the execution trace.
\end{itemize}
This design decouples the feedback mechanism from the scalar optimization objective, allowing the Analyst to pass structural and qualitative insights back to the optimizer.

\paragraph{Update Step (Generator).}
The Generator module $\mathcal{G}$ functions as a semantic evolutionary operator. It aggregates the current population $P_t$ and the set of feedback signals $\{\Delta_t^{(i)}\}_{i=1}^N$ to synthesize the next generation:
    \begin{equation}
        P_{t+1} \gets \mathcal{G}(P_t, \{\Delta_t^{(i)}\}_{i=1}^N).
    \end{equation}
Unlike standard genetic algorithms that rely on blind mutation or random crossover, $\mathcal{G}$ utilizes an LLM to perform \textbf{reasoned evolution}. By analyzing the feedback $\Delta$ across the population, the generator constructs new candidates that conceptually represent a directed jump in the search space. This process replaces stochastic genetic operators with a learned, context-aware optimization step that maximizes the population's expected improvement.

\subsection{Framework Instantiation}
The Forward, Backward, and Update modules form an iterative cycle that evolves the population $P_t$. The detailed procedure for transitioning from $P_t$ to $P_{t+1}$ is as follows:

\paragraph{Initialization.}
The process begins with empty feedback and a seed population $P_0$ generated via zero-shot prompting on the problem description $\mathcal{T}$, creating an initial baseline for the optimization process.

\paragraph{Step 1: Selection (Generator).} One or more subset(s) of parent heuristics $P_{parents} \subset P_t$ are sampled by a selection customized operator;

\paragraph{Step 2: Directed Generation (Generator).} For each parent set $P_{parents}$, the Generator $\mathcal{G}$ is prompted with the parent's code and their semantic gradient $\Delta_{parents}=\{\Delta_{t-1}^{(i)}|\theta^{(i)}\in P_{parents}\}$ to generate a new candidate $\theta'$,
\begin{equation}
   \theta' \sim \text{LLM}_{\mathcal G}(P_{parents}, \Delta_{parents}, \mathcal{T}), 
\end{equation}
formulating a batch of new candidates $P_{new} = \{\theta'_1, \dots\}$.

\paragraph{Step 3: Candidate Evaluation (Forward Pass).} Each new candidate $\theta'\in P_{new}$ is then injected into the algorithm skeleton and evaluated on the training set to get the execution traces $\mathcal{E}_{\theta'}$;

\paragraph{Step 4: Semantic Gradient Estimation (Backward Pass):} The analyst $\mathcal A$ then analyzes the execution traces $\mathcal E_{\theta'}$ and generate feedback $\{\Delta_t'\}$ for each heuristic $\theta'\in P_{new}$;
    
\paragraph{Step 5: Population Survival (Generator).}
The population is updated by selecting $N$ survivors from the union of the current population and the new candidates, $P_t \cup P_{new}$. This selection is driven by a survival function $\mathcal{S}$ that evaluates the comprehensive feedback of all candidates:
\begin{equation}
P_{t+1} = \mathcal{S}\left( { (\theta, \Delta_t^\theta) \mid \theta \in P_t \cup P_{new} } \right).
\end{equation}
This formulation allows the survival mechanism to consider not just scalar fitness, but also trade-offs in runtime or structural diversity encoded in $\Delta$.

The framework iterates steps 1-5 until a maximum budget of evaluations is reached. 

%% file: tex/06-method.tex
\subsection{Detailed Module Design}
We now highlight the key module-level design choices that distinguish our method from prior LLM-based heuristic search methods.

\paragraph{Forward Pass: Co-evolution of Heuristic Components} 
In our framework, we enable the joint optimization of multiple algorithmic components. Specifically, we target the Large Neighborhood Search (LNS) paradigm and genetic algorithm (GA) and co-evolve:
\begin{enumerate}
    \item \textbf{Constructive Heuristic ($\theta_{cons}$):} Responsible for generating high-quality initial solutions, where the later local search is more promising and landscape more smooth. Unlike traditional step-by-step construction, we only request the heuristic to output a solution, and allow this solution to later perform limited search. The constructive heuristic here is aimed to enable the solver begin in a promising region of the landscape.
    \item \textbf{Refinement Heuristic ($\theta_{ref}$):} Responsible for selecting neighborhoods (sub-problems) in LNS or selecting parent heuristic(s) in GA to re-optimize or mutate/crossover. This heuristic must identify local bottlenecks in the solution structure.
\end{enumerate}
Co-evolving these components allows the system to discover synergistic strategies (e.g., a constructive heuristic that leaves specific "gaps" which the refinement heuristic is particularly good at filling) that are inaccessible when evolving components in isolation.

\paragraph{Code-writing analyst in backward pass.} 
The backward pass provides critical guidance for heuristic synthesis. While a scalar fitness signal $f(\theta_t^{(i)})$ provides a global performance measure, its sparsity fails to characterize the specific structural failure of a heuristic. Conversely, purely textual reflection often lacks the precision required to guide code synthesis for complex combinatorial landscapes. 
To bridge this gap, we propose to use an executable 'Analyst' module, which interprets the backward pass as a feature discovery task. Leveraging its own analytical history, the Analyst synthesizes a new set of feature functions $\Psi:(\Phi, x)\mapsto \mathbb R$ at each call, which are designed to capture structural properties of the problem instances and corresponding solutions. These functions are optimized to distinguish high-quality solutions from poor ones based on specific attributes without computing the cost. 
Once the Analyst generates a set of features $\{\Psi_j\}_{j=1}^M$, the functions are executed across the training set $\mathcal{D}_{train}$ to produce a feature distribution for each heuristic $\theta_t^{(i)}$. The resulting structured feedback $\Delta_t^{(i)}$ is defined as a tuple of distributional statistics:
\begin{equation}
    \Delta_t^{(i)} = \left(f(\theta_t^{(i)}), \text{range}(\Psi), \mu_{\Psi}, \Psi_{best}, \Psi_{worst} \right)
\end{equation}
where $\mu_{\Psi}$ is the mean feature value across all (instance, solution) pairs, and $\Psi_{best}$/$\Psi_{worst}$ represent the feature signatures of the highest and lowest-performing solutions, respectively. 
When provided with these distributions, the Generator aims to generate heuristics that improve the features of poorly performing instances to the good ones, rather than performing a blind search.

\paragraph{Sampling and population survival in generator.}
To mitigate mode collapse, where the population converges prematurely to a single local minimum, the generator $\mathcal G$ employs a diversity-aware strategy for both parent selection and population survival. This design builds on prior observations that behavioral diversity is critical for effective heuristic evolution~\cite{dat2025hsevo}.
For each heuristic $\theta$, we maintain a fitness vector $\mathrm{v}_\theta=[f(\Phi_1,h_\theta(\Phi_1),\cdots,f(\Phi_K, h_\theta(\Phi_K))]$ representing its performance across the training instances. 
Rather than selecting parents based solely on average fitness, we compute the pairwise Euclidean distance between these vectors and prioritize pairs that are both high-performing and behaviorally diverse. This encourages the LLM to perform "semantic crossover" by merging distinct algorithmic strategies that succeed on different types of problem instances.
When updating population $P_t$ with newly generated candidates $\{\theta'\}$, we first preserve a subset of top-performing individuals to ensure the search remaining high quality. The remaining slots in $P_{t+1}$ are filled by randomly selecting candidates from $P_t\cup \{\theta'\}$ by a weighted criterion that balances fitness and marginal diversity. This ensures the new population maintain strong heuristics while continuing to explore a broad range of heuristic behaviors.

\paragraph{Relation to existing methods}
\input{table/method-comparison}
Existing LLM-based heuristic design methods can be interpreted as specific instantiations of our proposed modular framework, as summarized in~\cref{tab: method-comparsion}. 
In EoH~\cite{fei2024eoh}, the forward pass optimizes a single heuristic component, while the backward pass doesn't provide any explicit feedback (not even fitness value). The heuristic updates are driven by mutation/crossover over randomly selected parents under fixed prompts.
LLM-LNS~\cite{ye2025llmlns} augments this pipeline with scalar fitness value as feedback.
ReEvo~\cite{ye2024reevo} further introduces textual reflection to the backward pass and updates the heuristic based on crossover over the best-performing heuristics and randomly-selected parents.

By explicitly decoupling evaluation, feedback, and update mechanisms, our formulation exposes these design choices and enables systematic extensions of joint optimization of multiple heuristic components and richer feedback generation. 
While prompt evolution, as in LLM-LNS~\cite{ye2025llmlns}, is fully compatible with the generator module of our framework, we do not include it as a modular upgrade in this work to isolate the effects of feedback structure, heuristic component co-optimization, and population dynamics.

%% file: table/method-comparison.tex
\begin{table}[t]
\caption{\textbf{Comparison of representative methods with LaGO.} Existing approaches can be interpreted as restricted instantiations of our proposed forward–backward–update decomposition.}

    \label{tab: method-comparsion}
    \setlength{\tabcolsep}{10pt}
    \centering\small
        \begin{NiceTabular}{l|c|c|c}
        \toprule
        \textbf{Method} & \textbf{Forward} & \textbf{Backward} & \textbf{Update} \\
        \midrule
        \textbf{EoH} &  single & scalar & random\\
        \textbf{LLM-LNS} & single & scalar & random\\
        \textbf{ReEvo} & single & scalar + text & elitist\\
        \rowcolor{gray!15}
        \textbf{Ours} & joint & code-based & diversity\\
        \bottomrule
        \end{NiceTabular}
\end{table}

%% file: tex/07-exp.tex
\subsection{Experimental Settings}

\paragraph{Benchmarks and datasets.} 
We evaluate our framework on four diverse real-world combinatorial optimization problems selected from the HeuriGym benchmark~\cite{chen2025heurigym}. These tasks were chosen for their semantic diversity and complex, irregular constraints, ensuring that the evaluation measures the framework's reasoning capability rather than its retrieval of standard textbook algorithms (e.g., TSP heuristics). 
The problem set includes: pickup and delivery with time windows (PDPTW) and airline crew pairing from the logistics domain; technology mapping problem from electronic design automation; intra-operator parallelism scheduling (intra-op) from compiler applications. 

For each problem, we utilize the dataset of approximately $30$ instances. To rigorously assess generalization, we uniformly sample $50\%$ of instances from each data source to form the training set $\mathcal{D}_{train}$, and use the remaining instances as the test set $\mathcal{D}_{test}$. This small-data regime poses a significant challenge, requiring the framework to learn generalizable heuristic logic from limited feedback.

\input{table/main}

\paragraph{Baselines}
We compare our framework against representative LLM-based heuristic design methods, including HeuriGen~\cite{chen2025heurigym}, EoH~\cite{fei2024eoh}, ReEvo~\cite{ye2024reevo}, and LLM-LNS~\cite{ye2025llmlns}. For a comprehensive evaluation, we employ three distinct comparison protocols that reflect common practices in prior work:

\begin{itemize}
    \item \textbf{Improvement-Only (Primary Protocol):} 
    Most existing approaches focus on optimizing the \textit{refinement heuristic} while keeping the constructive heuristic fixed or randomly initialized. We adopt this as our primary comparison setting for consistency with the original evaluation procedures of EoH, ReEvo, and LLM-LNS.

    \item \textbf{Constructive-Only (Initialization Protocol):} 
    To examine the impact of initialization, we evaluate variants in which the baselines evolve only the \textit{constructive heuristic} and use a random improvement operator. These variants are denoted with the suffix \texttt{-I}.

    \item \textbf{End-to-End Generation:} 
    For HeuriGen, we follow its original protocol, where the LLM directly produces the complete executable code without a predefined algorithm skeleton.
\end{itemize}

These protocols allow us to isolate the performance contributions of different algorithmic components, contrasting the component-wise optimization of baselines against the \textbf{joint optimization} capability of our proposed framework.

\paragraph{Metrics.}
Following~\cite{chen2025heurigym}, we adopt the \textbf{Quality-Yield Index (QYI)} as the primary metric for evaluating performance. QYI provides a unified measure of a heuristic's robustness (feasibility) and optimality. It is defined as the harmonic mean of \texttt{Quality} and \texttt{Yield}:
\begin{equation}
    \texttt{QYI} = \dfrac{2 \cdot \texttt{Quality} \cdot \texttt{Yield}}{\texttt{Quality} + \texttt{Yield}}
\end{equation}
where the components are calculated as:
\begin{equation}
    \texttt{Quality} = \dfrac{1}{\hat K}\sum_{k=1}^{\hat K} \min\left(1, \dfrac {c^*_k}{c_k}\right), \quad \texttt{Yield} = \dfrac{\hat K}{K}
\end{equation}
Here, $K$ is the total number of instances, $\hat K$ is the number of instances for which a feasible solution was found, and $c_k$ and $c_k^*$ represent the costs of the LLM-generated solution and the expert or optimal solution for instance $k$, respectively.

While QYI is used for final evaluation, directly optimizing hard constraints can lead to sparse rewards (sparse feedback). To facilitate smoother gradient estimation during training, we relax hard constraints into soft penalties. For internal fitness computation, we treat all generated solutions as "valid" (i.e., setting effective $\texttt{Yield} \equiv 1$) but add a large penalty term $M$ to the cost $c_k$ for every constraint violation. This allows the framework to distinguish between "promising but invalid" heuristics and "completely broken" ones, providing a stronger signal to the Analyst module.
\begin{figure*}
    \centering
    \includegraphics[width=1\linewidth]{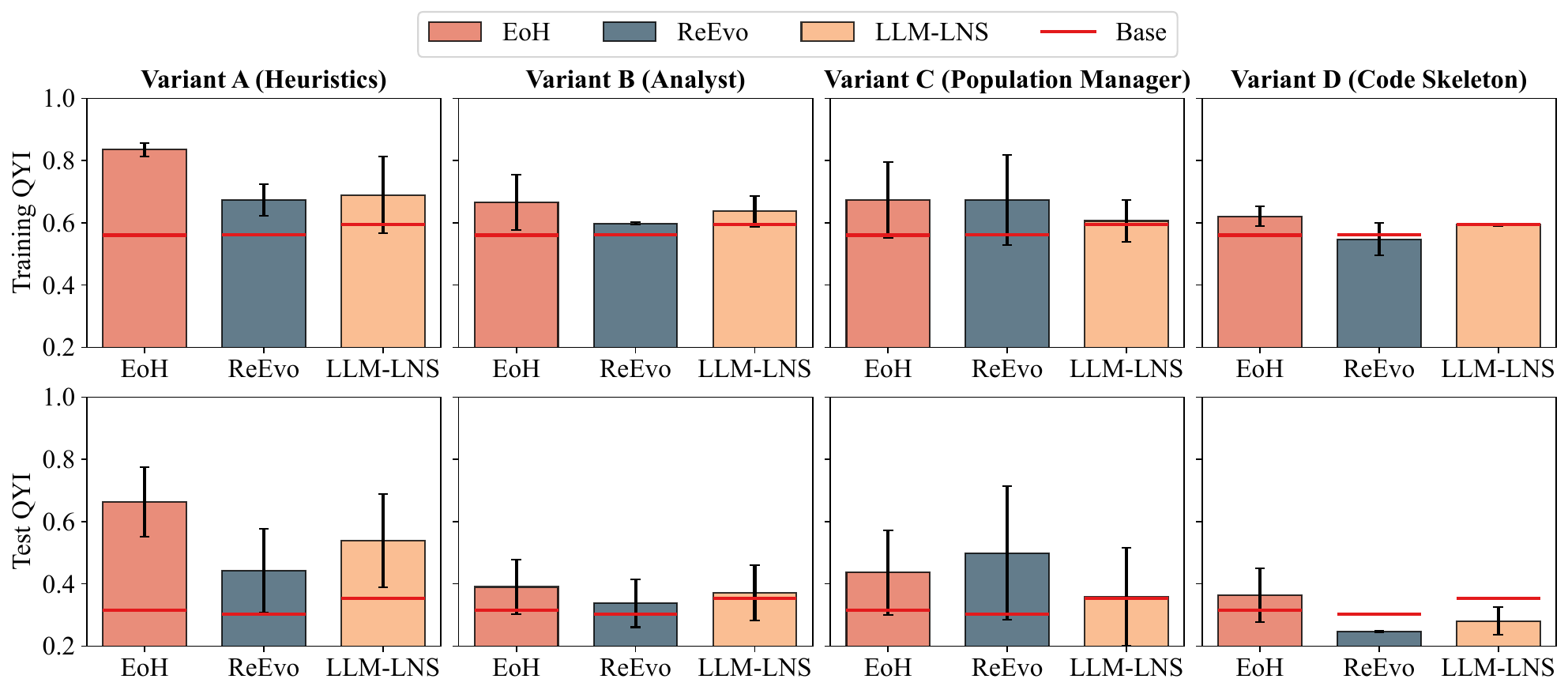}
    \caption{\textbf{Component analysis of framework modules on PDPTW} To isolate the performance gains contributed by specific components of our framework, we independently upgrade the baseline models with each proposed module, while keeping other modules the same as the original. Variant A introduces the co-evolution of constructive and refinement heuristics; variant B uses a code-writing analyst module; variant C applies a diversity-aware population management strategy; variant D adds the algorithm skeleton code to the generator prompt.}
    \label{fig:modular_ablation}
\end{figure*}

\subsection{Results}
We first report the performance of our full framework and all baselines across four combinatorial optimization domains in~\cref{tab: original-baseline} using \texttt{Gemini-3-Flash} \cite{team2025gemma} with temperature $T=1$ and \texttt{medium} reasoning effort. The performance is reported by averaging $3$ runs. Here, we highlight four key empirical findings.

\paragraph{End-to-End Generation is Ineffective for Constrained Heuristics.}
The HeuriGen baseline~\cite{chen2025heurigym}, which relies on end-to-end code generation, performs poorly across all domains and solves only a limited number of them. We also observed that the generated heuristics frequently violate problem constraints. This result indicates that direct program synthesis is insufficient for real-world combinatorial optimization with complex feasibility requirements.

\paragraph{Joint Optimization Outperforms Partial Optimization.}
Under the standard protocol where baselines optimize only improvement heuristics, our framework consistently achieves the strongest performance across all domains. On the highly constrained PDPTW and intra-operative scheduling tasks, our method achieves QYI scores of {0.882} and {0.783} on the training set, outperforming the strongest baseline by {+0.29} and {+0.27}, respectively. 
When baselines are instead configured to generate initial solutions, our method still maintains a performance advantage, achieving an average QYI gap of 0.09 compared to the best baseline. Although the relative gap is smaller than in the improvement-focused regime, the consistent advantage across both protocols indicates that optimizing only initialization or improvement in isolation is insufficient. 
Overall, these results demonstrate that jointly optimizing the full algorithmic pipeline yields substantially higher solution quality. The expanded scope provides greater expressivity, allowing LaGO to adapt to the training distribution while simultaneously exploring a broader, synergistic search space that partial optimization methods cannot access.

\paragraph{Generalization Capability.} 
Finally, we analyze the performance drop between training (seen instances) and testing (all instances). Baselines such as EoH suffer from an average QYI drop of 0.24 on the PDPTW test set. In contrast, our framework maintains a much tighter generalization gap of only 0.07. This suggests that the heuristics obtained from our framework transfer more reliably across problem instances than they overfit to specific training cases.

\subsection{Cross-Method Component Analysis}
To demonstrate that the performance gains are driven by our proposed modules, we conduct a cross-method component analysis on the PDPTW task. Instead of limiting the ablation study to the LaGO framework, we independently integrate each proposed module into three baseline methods (EoH, ReEvo, and LLM-LNS), while keeping all other components as originally defined. Specifically, we evaluate: (A) co-evolution of constructive and refinement heuristics, (B) code-writing analyst for structured feedback, (C) diversity-aware population management. We also evaluate another variant (D) of explicitly adding the skeleton algorithm to the generator. Results in~\cref{fig:modular_ablation} demonstrate that the LaGO framework's modular improvements are quite transferable.

\paragraph{Joint Optimization Boosts Cold-Start Method.} 
The introduction of co-evolving constructive and refinement heuristics (Variant A) yields a consistent performance boost across all methods, especially EoH, with a \texttt{QYI} score increase of $0.3$ over the baseline (red dashed line). This confirms that evolutionary methods lacking a strong initialization mechanism suffer from a cold start problem, in which they become stuck in a bad local minimum, especially when the fitness landscape is rugged and soft constraints are present. By optimizing the constructive heuristic to provide a valid starting point, Variant A enables the solver to finally converge to a feasible and better solution.

\paragraph{Analyst Complements Sparse Feedback.} 
Introducing the code-writing Analyst (Variant B) substantially improves performance for EoH and LLM-LNS, both of which rely primarily on scalar reward signals. The Analyst provides structured, instance-aware feedback that complements sparse fitness values and enables more targeted heuristic updates. In contrast, ReEvo shows only marginal gains, consistent with its existing textual reflection mechanism. This result suggests that the Analyst's primary value lies in upgrading methods with limited feedback channels rather than replacing existing reflection-based designs.

\paragraph{Population Diversity Enable Sustained Exploration.} 
Diversity-aware population management (Variant C) benefits all baselines, with the largest gains $+0.2$ observed for ReEvo. While ReEvo’s reflection mechanism enables strong local refinement, it also tends to induce mode collapse, repeatedly elaborating on a narrow set of high-performing heuristics. Our diversity-driven sampling forces the selection of distinct parents, encouraging exploration of distinct algorithmic structures. This result indicates that managing population diversity is critical for long-horizon exploration in LLM-driven heuristic search.

\paragraph{Role of the Algorithm Skeleton.} 
Finally, we evaluate the impact of explicitly providing an algorithm skeleton (Variant D). Interestingly, this modification yields negligible improvements for simpler EoH and slight deficits for advanced methods like ReEvo. This suggests that, for single-heuristic frameworks, enforcing a static code structure constrains the LLM's exploration of diverse logic. 
The limited benefit of Variant D for baselines confirms that LaGO’s performance gains do not stem from providing a code template, but from the other three components.

\subsection{Further Analysis}
\paragraph{Impact of Problem Complexity}
\input{table/tsp}
To examine the applicability of our framework across problem domains of different maturity, we evaluate it on the Traveling Salesman Problem (TSP)~\cite{junger1995tsp}, following the same training and test protocols as EoH. As shown in~\cref{tab: TSP}, while relative gains are modest compared to real-world tasks, our method remains competitive with state-of-the-art constructive baselines~\cite{voudouris1999glstsp}, achieving the best results on three of six instances and ranking second on two others.

We attribute the more modest gains to two factors. First, TSP is a highly mature and well-studied problem domain: decades of research have produced specialized heuristics, such as Lin–Kernighan–Helsgaun~\cite{lin1973lkh}, that already operate near-optimally, leaving limited room for further algorithmic improvement. In contrast, many real-world optimization problems lack such dominant, hand-engineered solutions, where a flexible heuristic design framework can be more impactful. Second, our framework is particularly effective with richly structured semantics. Real-world tasks often involve complex logical constraints (e.g., precedence relations or time windows) that allow rich diagnostic feedback. By comparison, TSP is governed by metric distance, offering fewer opportunities for leveraging semantic feedback. These results suggest that our framework is most valuable in complex, under-explored domains where domain-specific heuristics are not yet fully optimized.

\paragraph{Search Dynamics and Efficiency.}
\begin{figure}
    \centering
    \includegraphics[width=1\linewidth]{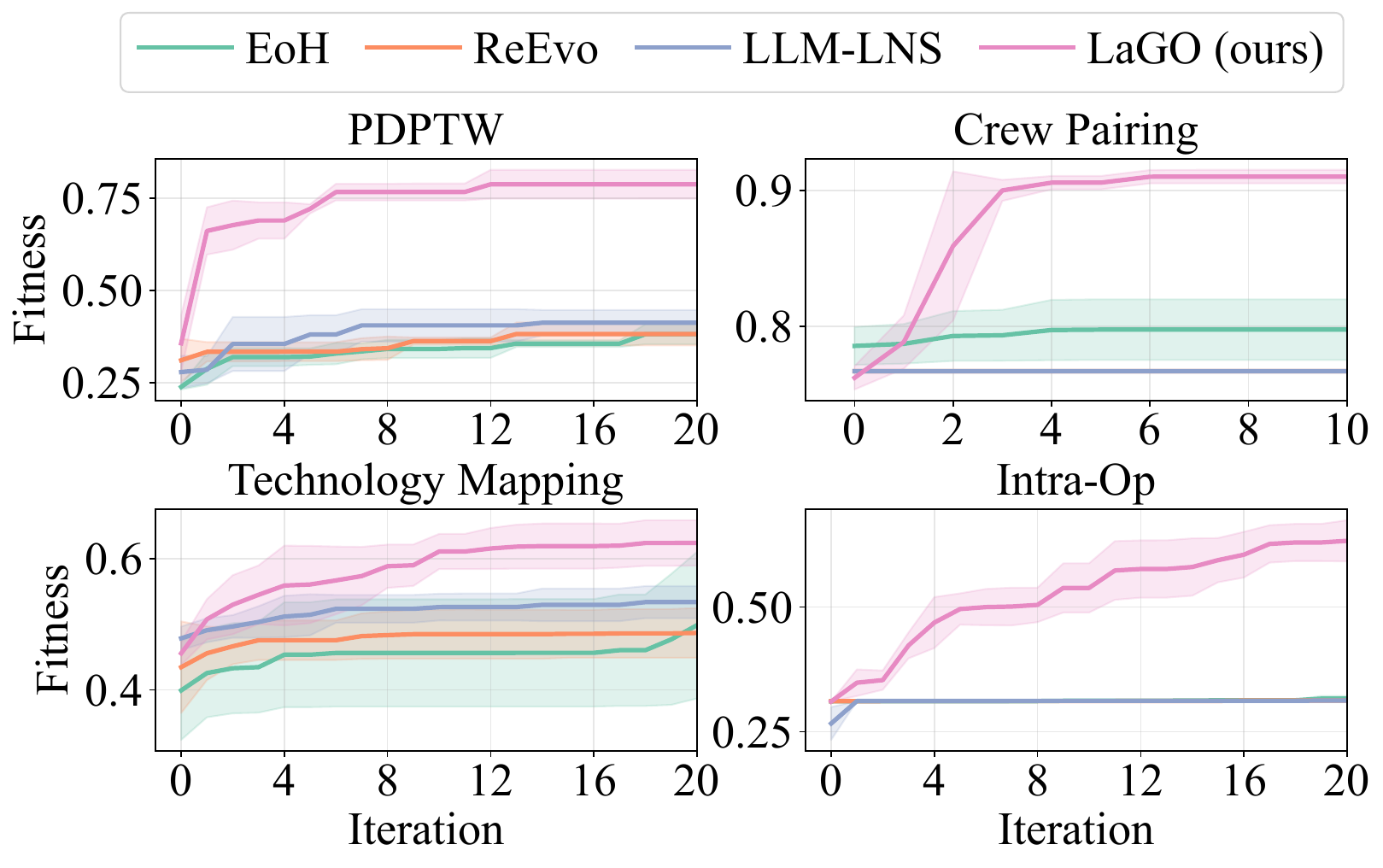}
    \caption{\textbf{Convergence curves during training.} Despite optimizing a larger joint search space, our diversity-aware sampling enables stable convergence to a better final fitness score than single-refinement baselines.}
    \label{fig:convergence}
\end{figure}
We visualize the evolution of heuristic performance on the training set in~\cref{fig:convergence}. 
A key challenge in our framework is the significantly expanded search space introduced by jointly optimizing constructive and refinement heuristics, which may impede search efficiency and convergence. However, we observe that our method not only converges stably but also achieves a higher final fitness score than baselines restricted to refinement-only search. 
We attribute this efficiency to the diversity-aware management strategy in our update module. By actively maintaining diversity in the population and leveraging it in heuristic generation, our optimizer mitigates premature convergence to suboptimal local optima and achieves a more effective exploration-exploitation balance throughout the evolutionary process.

%% file: table/main.tex
\begin{table*}[t]
\caption{\textbf{Comparison of our framework with LLM-based heuristic design methods}. Results are reported in terms of QYI (higher is better). Baseline methods include end-to-end generation, improvement-only optimization, and construction-focused settings. Our framework consistently achieves the best performance across domains, demonstrating the advantage of our proposed framework.
}\label{tab: original-baseline}
    \setlength{\tabcolsep}{3.5pt}
    \centering\small
        \begin{NiceTabular}{lcccccccc}
        \toprule
        \multirow{2}{15pt}{Method}  & \multicolumn{2}{c}{PDPTW} & \multicolumn{2}{c}{Crew Pairing} 
        & \multicolumn{2}{c}{Techno Mapping} & \multicolumn{2}{c}{Intra-op} \\
        \cmidrule(lr){2-3}\cmidrule(lr){4-5}\cmidrule(lr){6-7}\cmidrule(lr){8-9}
         & Train & Test & Train & Test & Train & Test & Train & Test  \\
        \midrule
        Heurigen& $0.000_{\pm0.000}$ &  $0.000_{\pm0.000}$
                  & $0.605_{\pm0.429}$ & $0.603_{\pm0.428}$
                  & $0.206_{\pm0.356}$ & $0.210_{\pm0.364}$
                  & $0.284_{\pm0.060}$ & $0.477_{\pm0.071}$ \\
        \midrule
        EoH         & $0.560_{\pm0.035}$ & $0.314_{\pm0.057}$
                    & $0.881_{\pm0.018}$ & $0.853_{\pm0.000}$
                    & $0.677_{\pm0.086}$ & $0.686_{\pm0.009}$
                    & $0.513_{\pm0.001}$ & $0.522_{\pm0.044}$\\
        ReEvo     & $0.561_{\pm0.037}$ & $0.303_{\pm0.153}$
                  & $0.868_{\pm0.000}$ & $0.853_{\pm0.000}$
                  & $0.635_{\pm0.037}$ & $0.696_{\pm0.025}$
                  & $0.513_{\pm0.001}$ & $0.548_{\pm0.000}$\\
        LLM-LNS      & $0.595_{\pm0.051}$ & $0.354_{\pm0.129}$
                     & $0.866_{\pm0.000}$ & $0.853_{\pm0.000}$
                     & $0.687_{\pm0.042}$ & $0.699_{\pm0.025}$
                     & $0.514_{\pm0.002}$ & $0.532_{\pm0.023}$  \\
        \midrule
        EoH-I  & $0.838_{\pm0.005}$ & $0.521_{\pm0.042}$
                  & $0.947_{\pm0.004}$ & $0.888_{\pm0.035}$
                  & $0.629_{\pm0.002}$ & $0.678_{\pm0.001}$
                  & $0.610_{\pm0.031}$ & $0.450_{\pm0.076}$ \\
        ReEvo-I & $0.859_{\pm0.042}$  & $0.636_{\pm0.065}$
                   & $0.912_{\pm0.038}$& $0.882_{\pm0.032}$
                   & $0.631_{\pm0.001}$ & $0.681_{\pm0.003}$
                   & $0.597_{\pm0.110}$ & $0.542_{\pm0.054}$\\
        LLM-LNS-I  & $0.836_{\pm0.002}$ & $0.487\pm_{0.032}$
                      & $0.936_{\pm0.035}$ & $0.916_{\pm0.026}$
                      & $0.629_{\pm0.002}$ & $0.682_{\pm0.002}$
                      & $0.560_{\pm0.031}$ & $0.515_{\pm 0.081}$  \\
        \midrule
        Ours   & $\mathbf{0.882}_{\pm0.031}$ & $\mathbf{0.808}_{\pm0.002}$
               & $\mathbf{0.953}_{\pm0.003}$ & $\mathbf{0.918}_{\pm0.006}$
               & $\mathbf{0.732}_{\pm0.061}$ & $\mathbf{0.713}_{\pm0.013}$
               & $\mathbf{0.783}_{\pm0.045}$ & $\mathbf{0.636}_{\pm0.060}$\\
        \bottomrule
        \end{NiceTabular}
\end{table*}

%% file: table/tsp.tex
\begin{table}[t]
\caption{\textbf{Traveling Salesman Problems Heuristic Comparison.} This table provides a sanity check of the proposed method on well-studied combinatorial problems. The best results are marked \textbf{bold}, the second-best results are marked by \underline{underline}.}\label{tab: TSP}
    \setlength{\tabcolsep}{3.2pt}
    \centering\small
        \begin{NiceTabular}{lcccccc}
        \toprule
        Method  & rd100 & pr124 & bier127 & kroA150 & u159 & kroB200 \\
        \midrule
        EoH   & 0.08 & \textbf{0.00} & \underline{0.49} & 1.12 & \textbf{0.00} & 2.39 \\
        ReEvo  & \textbf{0.00} & 0.09 & \textbf{0.41} & \textbf{0.00} & \textbf{0.00} & \textbf{0.90} \\
        LLM-LNS  & 0.01 & 0.08 & 0.62 & 1.35 & 0.32 & 1.92 \\
        \midrule
        Ours   & \textbf{0.00} & \textbf{0.00} & 1.06 & \underline{0.45} & \textbf{0.00} & \underline{1.19}\\
        \bottomrule
        \end{NiceTabular}
\end{table}

%% file: tex/09-conclusion.tex
In this work, we introduced a unified framework for large language model-based automated heuristic design. By reformulating the heuristic discovery process as an explicit optimization problem over the program space, we moved beyond ad hoc evolutionary pipelines to a principled Language-Guided Optimization paradigm. Decomposing the search loop into forward, backward, and update modules enabled us to systematically improve existing methods.
Extensive empirical evaluations on challenging real-world benchmarks, including PDPTW and crew pairing, demonstrate that our proposed instantiations yield significant performance gains over state-of-the-art baselines.

More broadly, our framework provides a foundation for future research in automated algorithm discovery.
In particular, it offers a common abstraction for reasoning about, comparing, and extending LLM-based AHD methods across domains. Potential directions include dynamic component selection within heuristic pipelines; alternative feedback modalities beyond code execution-based profiling; advanced update strategies beyond genetic programming.

%% file: appendix/related.tex
\paragraph{Code Generation with Execution Feedback}
Beyond the specific domain of heuristics, there is a broader literature on how LLMs can iteratively improve executable code through interaction with execution environments~\cite{wang2024surveyagent}. This paradigm has been explored in diverse settings, including debugging~\cite{chen2023teaching,tian2024debugbench}, algorithmic competition challenges~\cite{li2022competition, shi2024can}, and general planning problems~\cite{zhang2023planning,hao2025large}. 
A common theme in these approaches is the use of execution-based feedback, such as test outcomes, traces, or scalar rewards—to drive iterative refinement~\cite{liventsev2023fully,guo2024deepseek}.
This idea has proven effective in applications ranging from reward design for reinforcement learning~\cite{ma2023eureka, kwon2023reward} to algorithm self-improvement~\cite{zelikman2024self}.
However, despite their success, most existing methods treat the optimization loop as a tightly coupled, end-to-end process, where evaluation, diagnosis, and code generation are intertwined. 
In the specific context of automated heuristic design for CO, where solution quality depends on subtle structural interactions and constraint satisfaction, this design makes it difficult to reason about the role of individual components, compare methods in a principled manner, or systematically extend them to more structured domains such as combinatorial optimization.

%% file: appendix/heurigym.tex
There are nine real-world combinatorial optimization problems from Heurigym~\cite{chen2025heurigym} that were excluded from the main paper. In this paper, we include four problems from the benchmark and exclude the remaining five problems. The main criterion is the quality of the greedy algorithm's solution. Simple greedy algorithms can find optimal or near-optimal solutions for all the test cases of Operator Scheduling, E-Graph Extraction, Protein Sequence Design, Global Routing, and Mendelian Error Detection. The problems, hence, are not challenging enough to evaluate the performance of LLM-based heuristic evolution pipelines with the latest LLMs. Here we introduce the problems included in \cref{tab: original-baseline}.

\subsection{Pickup and Delivery Problem with Time Windows}
The Pickup and Delivery Problem with Time Windows (PDPTW) \cite{dumas1991pickup} is a generalization of the Capacitated Vehicle Routing Problem (CVRP), in which a fleet of vehicles must serve a set of transportation requests, each consisting of a pair of pickup and delivery locations. The core challenge is to construct a set of routes that minimizes travel cost (distance or time) and fleet size while strictly adhering to pairing and precedence constraints—specifically, that a request's pickup must precede its delivery on the same vehicle. Additionally, service at every location must commence within a specified time window, and the vehicle's load must never exceed its capacity. This problem is widely used to benchmark solvers on their ability to handle precedence-constrained routing and tight temporal feasibility checks, which are central to applications like on-demand logistics, paratransit services, and autonomous ride-sharing.

\subsection{Airline Crew Pairing}
The Airline Crew Pairing Problem (CPP)~\cite{andersson1998crew} is formulated as a cost-minimization problem over a set of flight legs, modeled as a large-scale Set Partitioning Problem (SPP) or Set Covering Problem (SCP). The objective is to construct a set of "pairings", e.g., sequences of flight legs starting and ending at the same crew base, to cover every flight leg exactly once. This problem is characterized by highly constrained search spaces due to complex validity regulations, such as FAA safety rules on maximum flying time and minimum rest periods. The total costs are a complex function of factors such as wages, hotel accommodations, and deadhead travel. We utilize CPP instances to evaluate the solver's ability to handle high-dimensional combinatorial constraints and non-convex objective landscapes.

\subsection{Technology Mapping}
Technology Mapping \cite{chen2004daomap} is a fundamental step in the Electronic Design Automation (EDA) logic synthesis flow, responsible for binding a technology-independent Boolean network (typically an And-Inverter Graph) to a specific library of physical gates. The problem is formally defined as a graph covering problem: finding a valid cover of the subject graph using a set of $K$-input subgraphs. The optimization goal is to minimize the total number of subgraphs. We utilize CPP instances to evaluate the solver's ability to handle high-dimensional combinatorial constraints and non-convex objective landscapes. This problem serves as a rigorous benchmark for combinatorial optimization under large input size.

\subsection{Intra-Operator Parallelism}
The Intra-Operator Parallelism (IOPDDL)~\cite{moffitt2025asplos} is a combinatorial optimization challenge introduced in the ASPLOS’25 contest track, focusing on the efficient distribution of large-scale deep learning models across multiple hardware accelerators. The problem is formulated on a large computational graph, where nodes represent operations and edges represent tensor dependencies. For each node, the solver must select a single execution strategy from a set of discrete candidates, each defined by specific computational costs and memory requirements. The objective is to minimize the aggregate cost—comprising both operation execution (node cost) and communication overhead (edge cost)—while strictly adhering to peak memory usage constraints on every device. Due to the diversity of model topologies and the scale of operations involved, this problem serves as a rigorous benchmark for constrained graph optimization, where leading solutions currently rely on specialized meta-heuristic strategies.

%% file: appendix/more-results.tex
\subsection{Experiment Detail}
In the experiments on all tested domains (including TSP), we run the automatic heuristic design for $20$ iterations, except for $10$ in crew pairing, as it's simple enough to converge at an early stage. And the population size is $10$, except for $4$ in crew pairing.

\subsection{Additional Experiment}
\input{table/PDPTW-different-model}

We conduct an additional experiment on PDPTW using a previous-generation model, \texttt{GPT-4.1-mini}. The experimental results, summarized in Table \ref{tab: PDPTW-detailed}, demonstrate that our framework is not tailored to a specific LLM architecture but rather generalizes effectively across different backends. While baseline methods exhibit inconsistent behavior when transitioning from \texttt{GPT-4.1-mini} to \texttt{Gemini-3-Flash}, e.g., ReEvo suffers a performance degradation ($\Delta = -0.030$), our approach maintains robust superiority on both models. This confirms that the proposed method captures the intrinsic structure of the PDPTW problem, rather than exploiting the idiosyncrasies of a single language model. Furthermore, our framework benefits most significantly from the enhanced reasoning capabilities of \texttt{Gemini-3-Flash}, achieving the largest performance gain on the test set ($+0.171$). This positive scaling indicates that, unlike prior heuristics (e.g., LLM-LNS and EoH), which see diminishing or marginal returns, our method is uniquely positioned to leverage advancements in next-generation foundation models.

%% file: table/PDPTW-different-model.tex
\begin{table}[t]
\caption{\textbf{PDPTW Performance Evaluation with \texttt{GPT4.1-mini}.}}\label{tab: PDPTW-detailed}
    \setlength{\tabcolsep}{7pt}
    \centering\small
        \begin{NiceTabular}{lcccc}
        \toprule
        Method  & GPT train & Switch Gain & GPTtest & Switch Gain \\
        \midrule
        EoH    & $0.469$ & $0.091$ & $0.244$ & $0.070$\\
        ReEvo  & $0.467$ & $0.094$ & $0.333$ & $-0.030$ \\
        LLM-LNS    & $0.533$ & $0.063$ & $0.246$ & $0.108$ \\
        \midrule
        Ours   & $0.736$ & $0.146$ & $0.637$ & $0.171$ \\
        \bottomrule
        \end{NiceTabular}
\end{table}

%% file: appendix/prompt.tex
We present the prompt used in the framework for analyst LLM and generator LLM, respectively.
\subsection{Analyst}
    
    \begin{tcolorbox}[colback=gray!3,       %
        colframe=gray!30,     %
        coltitle=black,       %
        fonttitle=\bfseries,
        sharp corners,        %
        boxrule=0.5pt, title=System Prompt]
        You are a world-class optimization expert specializing in metaheuristics and LNS. Your task is to design a suite of diagnostic features that characterize both the difficulty of a problem instance and the quality/bottlenecks of a current solution. \\

        \textbf{Objectives}:
        \begin{enumerate}
            \item Discriminative Power: Features must help distinguish why certain instances are harder than others and why specific solutions are suboptimal. Please also be noted that different instances may have significant difference in scale and absolute cost values, which doesn't necessarily reflect the performance.
            \item Feasibility Analysis: Capture violations, which are represented as soft constraint costs.
            \item Diversity: Include spatial, temporal, and other structural metrics if applicable.
        \end{enumerate}
        
        \textbf{Implementation Requirements}:
        \begin{itemize}
            \item Input Argument: Every function must accept exactly the arguments as provided in the example.
            \item Output: Each function must return a single `float`.
            \item Aggregation: You must provide a list named `feature\_func\_list` containing all your function objects.
            \item Environment: Python 3.12. Use `numpy` for efficiency. Include all necessary imports.
            \item Strictness: Output ONLY the Python code. No explanations, no markdown code blocks, no comments outside the code.
        \end{itemize}

        \textbf{Problem Information}: \{problem\_description\}\\
        \textbf{Template analyst code}: \{template\_analyst\_code\}

    \end{tcolorbox}

    \begin{tcolorbox}[colback=gray!3,       %
        colframe=gray!30,     %
        coltitle=black,       %
        fonttitle=\bfseries,
        sharp corners,        %
        boxrule=0.5pt, title=User Prompt]
        The following features were used in the previous iteration:
            \{history\_feature\_code\}\\

        Your analysis for currently best individual is \{current\_best\_heuristic\_feature\},\\
        Whereas the solution costs are as follows: \{current\_best\_heuristic\_training\_cost\}\\

        \textbf{(if no improved in last iteration)}\\
        The previous iteration did NOT show improvement in heuristic performance. This suggests the current features might not be capturing the critical aspects of the problem needed for better decision-making. Please significantly modify or add new features that provide deeper insights into the solution state.\\
        
        \textbf{(if improved in last iteration)}\\
        The previous iteration showed improvement in heuristic performance. You can choose to refine the existing features or add complementary ones to further capture the problem structure.

    Please provide several feature functions that can characterize the state of a solution for this problem or the critical aspects of the problem instance. End your response with a list named `feature\_func\_list` containing the function objects.

    \end{tcolorbox}

    \begin{tcolorbox}[colback=gray!3,       %
        colframe=gray!30,     %
        coltitle=black,       %
        fonttitle=\bfseries,
        sharp corners,        %
        boxrule=0.5pt, title=User Prompt (Error Fixing)]
        Your previous feature code encountered errors during execution. Please identify and fix the issues in the feature functions to ensure they run correctly and provide meaningful analysis.
        Use same formatting as provided feature code and modify only the necessary parts to fix the errors.\\

        \textbf{Previous Feature Code}:\{past\_feature\_code\}\\
        \textbf{The errors are as follows}:\{best\_individual\_analysis\_traceback\_errors\}

    \end{tcolorbox}

\subsection{Generator}

\begin{tcolorbox}[colback=gray!3,       %
        colframe=gray!30,     %
        coltitle=black,       %
        fonttitle=\bfseries,
        sharp corners,        %
        boxrule=0.5pt, title=System Prompt]
        The task can be solved step-by-step by starting from a constructive initial solution and iteratively selecting a subset of decision variables to relax and re-optimize. In each step, most decision variables are fixed to their values in the current solution, and only a small subset is allowed to change. You need to score all the decision variables based on the information I give you, and I will choose the decision variables with high scores as neighborhood selection. To avoid getting stuck in local optima, the choice of the subset can incorporate a degree of randomness.\\

        + You need to come up with a heuristic function which score a state. A good practice is to analyze the cost function and current solution to find what contributes most to the cost. Think systematically and creatively. Also I suggest the score function to be weighted sum. Note that the cost function actually soft constraints with very heavy penalty. You should pay more attention to these than the normal cost.\\
        
        + You also need to provide an initial solution function, which should be a constructive heuristic. The initial solution itself does not necessarily perform well, but it needs to provide a warmstart for neighborhood search to get close to a good local minimum. You also need to pay attention to the synergy between the initial solution and heuristics.\\
        
        + You should be noted that the generated solution should be general for out-of-distribution data, which may vary in scale and characteristic of numbers. So don't just exploit what you know, try to find different strategies which all work good on training sets.\\
        
        + A problem template is provided below. You need to implement both the 'heuristic' and '\_init\_solution' function. Do NOT modify the function signature including the data types of the input arguments. Make sure you have imported necessary libraries and modules.\\

        Template Heuristic Code: \{template\_heuristic\}
    \end{tcolorbox}

    \begin{tcolorbox}[colback=gray!3,       %
        colframe=gray!30,     %
        coltitle=black,       %
        fonttitle=\bfseries,
        sharp corners,        %
        boxrule=0.5pt, title=User Prompt (Operator i1 - initialization)]
        "Your task is to initialize a new heuristic algorithm for scoring decision variables. "
        "Refer to the format of a trivial design provided. Be very creative."\\
        
        "Step 1: Briefly describe your new algorithm and its main steps in one sentence. This description MUST be inside braces \{\} and commented by """ """ as in Python."\\
        
        "Step 2: Implement the algorithm in Python. Ensure all necessary imports are included, and the function signature matches the template."

    \end{tcolorbox}

    \begin{tcolorbox}[colback=gray!3,       %
        colframe=gray!30,     %
        coltitle=black,       %
        fonttitle=\bfseries,
        sharp corners,        %
        boxrule=0.5pt, title=User Prompt (Operator e1 - crossover)]
        "I have two existing algorithms with their codes and performance features as follows:"\\
        
        "Algorithm 1 (better):"\\
        
        f"\{Individual 1\}"\\
        
        "Algorithm 2 (worse):"\\
        
        f"\{Individual\}"\\
        
        "Please help me create a new algorithm that is different from the given ones but can be motivated from them."
        "Your task is to synthesize a new algorithm by merging the most effective components as hints and design patterns from both."\\
        
        "Step 1: Briefly describe the new synthesized algorithm and its main steps in one sentence. This description MUST be inside braces {} and commented by """ """ as in python."\\
        
        "Step 2: Implement the new algorithm in Python. Ensure all necessary imports are included and the function signature matches the template."

    \end{tcolorbox}

    \begin{tcolorbox}[colback=gray!3,       %
        colframe=gray!30,     %
        coltitle=black,       %
        fonttitle=\bfseries,
        sharp corners,        %
        boxrule=0.5pt, title=User Prompt (Operator e2 - crossover)]
        "I have two existing algorithms with their codes and performance features as follows:"
        
        "Algorithm 1 (better):"\\
        
        f"\{Individual 1\}"\\
        
        "Algorithm 2 (worse):"\\
        
        f"\{Individual 2\}"\\
        
        "Your task is to evolve a new algorithm by identifying the differences in their performance and logic. "
        "Determine which algorithm performs better and why. Then, identify a limitation in the better algorithm that the other might address, "
        "or propose a novel mechanism inspired by their differences to overcome current bottlenecks."\\
        
        "Step 1: Briefly describe the new evolved algorithm and its main steps in one sentence. This description MUST be inside braces {} and commented by """ """ as in python."\\
        
        "Step 2: Implement the new algorithm in Python. Ensure all necessary imports are included and the function signature matches the template."

    \end{tcolorbox}

    \begin{tcolorbox}[colback=gray!3,       %
        colframe=gray!30,     %
        coltitle=black,       %
        fonttitle=\bfseries,
        sharp corners,        %
        boxrule=0.5pt, title=User Prompt (Operator m1 - mutation)]
        "I have one existing algorithm with its code and performance features as follows:"\\
        
        f"\{Individual\}"\\
        
        "Your task is to 'mutate' this algorithm to create a new variant. "
        "Analyze the current scoring logic and identify one core component or assumption that can be improved, "
        "diversified, or replaced with a different mathematical approach to explore a new area of the search space."\\
        
        "Step 1: Briefly describe the new mutated algorithm and its main steps in one sentence. This description MUST be inside braces {} and commented by """ """ as in python."\\
        
        "Step 2: Implement the new algorithm in Python. Ensure all necessary imports are included and the function signature matches the template."

    \end{tcolorbox}
    
    \subsection{Format of Individual Heuristic}
    \begin{tcolorbox}[colback=gray!3,       %
        colframe=gray!30,     %
        coltitle=black,       %
        fonttitle=\bfseries,
        sharp corners,        %
        boxrule=0.5pt, title=Individual]

        Response at iteration \{iteration\}

        code=\{code\}

        text\_description="""\{text description\}"""

        avg\_objective \{avg objective\}

        error\_msg='\{error msg\}'
        
        Performance Summary:
        
        - \{feature 1\}: Avg=\{avg feature 1 value\}, Range=[minimum feature 1 value, maximum feature 1 value]"\\
        - \{feature 2\}: Avg=\{avg feature 2 value\}, Range=[minimum feature 2 value, maximum feature 2 value]"\\
        $\cdots$\\
        - \{feature N\}: Avg=\{avg feature N value\}, Range=[minimum feature N value, maximum feature N value]"\\
        
        - Worst Instance (\{instance\}): Fitness=\{instance fitness\}, Features=\{instance feature values\}"

        - Best Instance (\{instance\}): Fitness=\{instance fitness\}, Features=\{instance feature values\}"
    \end{tcolorbox}